\documentclass[preprint,12pt]{elsarticle}

\usepackage{amssymb}
\usepackage{amsmath}
\usepackage{url}
\usepackage{float}
\usepackage{subcaption}
\usepackage{algorithm}
\usepackage{algorithmic}
\usepackage{multirow}
\usepackage{arydshln}

\journal{Applied Soft Computing}

\begin{document}

\begin{frontmatter}



\title{MemLoss: Enhancing Adversarial Training with Recycling Adversarial Examples}


\author[a]{Soroush Mahdi}
\ead{soroushm@aut.ac.ir}

\author[a]{Maryam Amirmazlaghani\corref{cor1}}
\ead{mazlaghani@aut.ac.ir}

\author[a]{Saeed Saravani}
\ead{s.saravani@aut.ac.ir}

\author[b]{Zahra Dehghanian}
\ead{z.dehghanian@aut.ac.ir}

\cortext[cor1]{Corresponding author: Maryam Amirmazlaghani\\
Department of Computer Engineering, Amirkabir University of Technology (Tehran Polytechnic), Iran\\
Tel: (+98) 021 6454-2704\\
Email: mazlaghani@aut.ac.ir}

\affiliation[a]{organization={Department of Computer Engineering, Amirkabir University of Technology (Tehran Polytechnic)},
            city={Tehran},
            country={Iran}}

\affiliation[b]{organization={Department of Computer Engineering, Sharif University of Technology},
            city={Tehran},
            country={Iran}}

\begin{abstract}
In this paper, we propose a new approach called MemLoss to improve the adversarial training of machine learning models. MemLoss leverages previously generated adversarial examples, referred to as 'Memory Adversarial Examples,' to enhance model robustness and accuracy without compromising performance on clean data. By using these examples across training epochs, MemLoss provides a balanced improvement in both natural accuracy and adversarial robustness. Experimental results on multiple datasets, including CIFAR-10, demonstrate that our method achieves better accuracy compared to existing adversarial training methods while maintaining strong robustness against attacks.
\end{abstract}

\begin{keyword}
Adversarial Robustness, Adversarial Attacks, Adversarial Training
\end{keyword}

\end{frontmatter}



\section{Introduction}
\label{sec:Introduction}

Adversarial attacks are designed to mislead machine learning (ML) models by feeding them manipulated inputs to undermine their effectiveness. Szegedy et al. \cite{szegedy2013intriguing} were the first to demonstrate that small, imperceptible perturbations to input data can lead neural networks to make incorrect predictions with high confidence. This discovery exposed a significant vulnerability in machine learning models and introduced the concept of adversarial attacks.

In recent years, the vulnerability of deep learning models to adversarial attacks has driven significant research into improving model robustness \cite{szegedy2013intriguing, goodfellow2014explaining}. Adversarial training, widely regarded as the most prominent defense against adversarial machine learning (AML) attacks, enhances model robustness by incorporating both benign and adversarial examples into the training process \cite{bountakas2023defense}. However, it often leads to reduced accuracy on clean data \cite{rade2022reducing}. This trade-off between robustness and accuracy presents a fundamental challenge:

\begin{quote}
\textit{``How can we improve a model's robustness against adversarial attacks without compromising its performance on clean data?''}
\end{quote}

To address the accuracy-robustness trade-off inherent in adversarial training, several refinements have been proposed to enhance the standard approach. Zhang et al. (2019) introduced the TRADES framework, which formalized the trade-off between robustness and accuracy by incorporating a regularization term that balances these objectives \cite{zhang2019theoretically}. Wang et al. (2020) focused on improving adversarial robustness by revisiting misclassified examples \cite{wang2019improving}, while Rice et al. (2020) highlighted the issue of overfitting in adversarially robust models \cite{rice2020overfitting}. Wu et al. (2020) proposed using adversarial weight perturbation to improve generalization \cite{wu2020adversarial}, and Zhang et al. (2021) introduced a geometry-aware approach to adversarial training that reweights instances based on their sensitivity to adversarial perturbations \cite{zhang2021geometry}. Additionally, the HAT method (Reducing Excessive Margin to Achieve a Better Accuracy vs. Robustness Trade-off) was proposed to minimize excessive margins in the decision boundary, thereby achieving a better balance between accuracy and robustness \cite{rade2022reducing}. Sankaranarayanan et al. introduced a regularization technique that enhances adversarial robustness by perturbing intermediate layers of DNNs \cite{sankaranarayanan2018regularizing}. All the aforementioned works update the decision boundary using adversarial examples generated in the current iteration, neglecting those produced in previous iterations. Consequently, the influence of current adversarial samples on the decision boundary may inadvertently increase the model's vulnerability to earlier adversarial examples.

In this paper, we propose a new method called \textbf{MemLoss}, which leverages \textit{Memory Adversarial Examples} to address this challenge. Memory Adversarial Examples are adversarial inputs generated in earlier epochs of training, and the concept of 'memory' in adversarial training signifies the model's ability to retain and utilize these previously encountered adversarial examples to enhance its robustness. By incorporating memory into adversarial training, MemLoss aims to maintain and even improve clean data accuracy while strengthening defenses against adversarial attacks.

From another point of view, previous research has shown that adding more data can significantly boost model performance \cite{carmon2019unlabeled}. However, relying on additional external datasets is not always feasible, especially when comparing models that do not use extra data. This situation leads us to look for ways to increase data without using outside sources, and we propose using Memory Adversarial Examples as a new source of data. This raises another critical question:

\begin{quote}
\textit{How should we use this information in the adversarial training process?}
\end{quote}

MemLoss leverages previously generated adversarial examples to establish a more effective and consistent training regimen. Unlike conventional adversarial training techniques, which discard adversarial examples after each epoch, MemLoss retains these examples throughout the training process. This retention allows the model to build upon past learning, thereby enhancing its overall robustness and accuracy.

MemLoss can easily integrate with any adversarial training framework, demonstrating its versatility. For example, when combined with the well-established TRADES framework \cite{zhang2019theoretically}, MemLoss significantly enhances its performance. Furthermore, we demonstrate that incorporating MemLoss not only improves TRADES but also augments the performance of HAT \cite{rade2022reducing}. These results highlight the broad applicability and effectiveness of our approach across various adversarial training methodologies.

The contributions of this paper are as follows:
\begin{itemize}
    \item We introduce MemLoss, a novel method that uses Memory Adversarial Examples to improve both the robustness and accuracy of machine learning models.
    \item We combine our method with two distinct adversarial training frameworks and demonstrate significant improvements in both clean and robust accuracy.
    \item We provide comprehensive experimental results on multiple datasets, including CIFAR-10 \cite{krizhevsky2009learning}, CIFAR-100 \cite{krizhevsky2009learning}, and SVHN \cite{netzer2011reading}. Our findings show that MemLoss achieves superior accuracy compared to existing adversarial training methods while maintaining strong robustness.
    \item We analyze the role of memory in adversarial training, demonstrating how it enhances training stability and overall model performance.
\end{itemize}

This paper is structured as follows. In Section \ref{sec:Related Work}, we review the existing literature relevant to adversarial machine learning, highlighting advancements and gaps. Section \ref{sec:Adversarial Training With Memory} introduces our novel approach, detailing its underlying principles and innovative aspects. In Section \ref{sec:Experiments}, we evaluate the efficacy of our method through extensive experiments, comparing its performance against state-of-the-art models. Finally, in Section \ref{sec:Conclusion}, we summarize our findings and discuss the implications and potential future directions of this research.

\section{Related Work}\label{sec:Related Work}
Adversarial examples are maliciously crafted inputs designed to fool deep neural networks into making incorrect predictions \cite{szegedy2013intriguing, goodfellow2014explaining}. Since the discovery of adversarial examples, many methods have been introduced for generating these examples, termed adversarial attacks \cite{mkadry2018towards, carlini2017towards}. This has led researchers to develop new defense strategies to protect against these threats. Adversarial training is one of the most recognized defenses against adversarial attacks. In this technique, the model is trained on adversarial examples (generated in the current epoch) to enhance its robustness against such attacks \cite{mkadry2018towards}.

In technical terms, let $x^i$ and $y^i$ be a normal example and its label in a dataset $S = \{(x^i, y^i)\}_{i=1}^{n}$, where $x^i$ is an input and $y^i$ is its label from 0 to $C-1$. We use $B_{\epsilon}[x^i]$ to describe all the inputs close to $x^i$, within a small range $\epsilon$, according to the $l_{\infty}$ norm. A deep neural network (DNN), defined by parameters $\theta$, acts as a function $f_{\theta}(\cdot): X \rightarrow \mathbb{R}^C$, which $X \subseteq \mathbb{R}^d$ is the input space, simplifying to $f(\cdot)$ for ease. To create an adversarial example, the goal is to adjust the input within its close range to maximize the error in the model's prediction or try to change the model prediction with minimum change in the input.

One of the most notable methods for adversarial training (AT) is Projected Gradient Descent (PGD) adversarial training \cite{mkadry2018towards}. This approach enhances model resilience by systematically incorporating adversarial examples into the training process.
PGD operates by generating adversarial examples through a series of iterative updates. It does this by leveraging the gradient of a specified loss function $\ell$ (for instance, the cross-entropy loss $CE(\cdot, y)$) in relation to the model's inputs. The process begins with an initial input $x'_0$, which is a sample from the training set. The updates are performed using the following equation:

\begin{equation}
   x'_{t+1} = \Pi_{B_{\epsilon}[x]}(x'_t + \alpha \cdot \text{sign}(\nabla_{x'_t}CE(f(x'_t), y)))
\end{equation}

In this equation, $x'_t$ denotes the adversarial example at iteration $t$ of the optimization process, and the function $\Pi_{B_{\epsilon}[x]}(\cdot)$ projects the modified input back into the $\epsilon$-bounded $l_{\infty}$ neighborhood of the original benign input $x$. The purpose of this projection function is to ensure that the generated adversarial examples remain within a defined boundary around the original input. By using these generated adversarial examples instead of clean inputs for training, Adversarial Training (AT) improves the model's robustness against adversarial attacks.
In adversarial machine learning, a key challenge is the trade-off between robust accuracy and clean accuracy. Clean accuracy measures a model's performance on unperturbed inputs, and Robust accuracy measures the model's performance on adversarial attacks. While incorporating adversarial examples during training can enhance robustness, it may also reduce clean accuracy, as the model might become overly specialized to those adversarial inputs, compromising its performance on clean data.

\textbf{TRADES} \cite{zhang2019theoretically} introduced a trade-off between robustness and accuracy by incorporating a Kullback-Leibler (KL) divergence term to balance clean accuracy and adversarial robustness. The TRADES loss formulation can be expressed as:
\begin{equation}
    \begin{aligned}
    \mathcal{L}_{\text{TRADES}}(\theta, x, y) = & \, \ell(f_{\theta}(x), y) \\
    & \, + \beta \cdot  KL(f_{\theta}(x) || f_{\theta}(x'))
    \end{aligned}
    \label{trades}
\end{equation}

In this equation, $\mathcal{L}_{\text{TRADES}}$ represents the TRADES loss, $\theta$ denotes the model parameters, $x$ represents the input example, $y$ is its corresponding label, $\ell$ stands for the standard classification loss (e.g., cross-entropy), $\beta$ is a regularization hyperparameter which controls the trade-off between accuracy and robustness, and $x'$ represents the adversarial example generated from $x$.

\textbf{Misclassification-Aware Robust Training (MART)} \cite{wang2019improving} builds on the concept of adversarial training by placing greater emphasis on misclassified examples during the training process. This approach aims to improve adversarial robustness by specifically addressing instances where the model struggles, thus enabling it to learn more effectively from its mistakes. Like PGD adversarial training, MART sacrifices clean accuracy to improve robustness.

\textbf{Helper-based Adversarial Training (HAT)} \cite{rade2022reducing} addresses the robustness-accuracy trade-off by proposing a method to reduce the adverse impact of adversarial training on clean data accuracy. It achieves this by minimizing the excessive margin between the decision boundaries in the model. Minimizing the excessive margin between decision boundaries means reducing the unnecessarily large gaps between classes created during adversarial training. By keeping the boundaries tighter, the model maintains robustness against attacks without sacrificing accuracy on clean data, striking a better balance between the two.

TRADES and HAT aimed to enhance model robustness without sacrificing accuracy, seeking to improve both clean accuracy and robust accuracy. These methods typically rely on adversarial examples generated in each iteration to adjust the decision boundary. However, they only consider the current adversarial example in the adjustment process. A more effective approach would involve updating the decision boundary by accounting for both the current adversarial example and those generated in earlier epochs, leading to better boundary adjustments.

\section{Adversarial Training With Memory}\label{sec:Adversarial Training With Memory}
In typical adversarial training, each epoch involves generating new adversarial examples $x'$ based on the current model parameters $\theta$. These examples are used for that specific epoch and then discarded. In the next epoch, the process repeats, using newly generated adversarial examples without retaining any from prior epochs.

Our proposed method, MemLoss, takes a different approach by reusing adversarial examples generated in earlier epochs, which we call Memory Adversarial Examples. This memory-based strategy enhances both robustness and clean accuracy by incorporating adversarial directions learned in previous epochs, creating a more stable and effective training process. Unlike conventional methods that discard adversarial examples after each epoch, MemLoss preserves them, allowing the model to benefit from accumulated knowledge throughout training. By utilizing previous adversarial examples, we ensure that adjustments to the decision boundary based on current adversarial examples do not make the model vulnerable to earlier adversarial examples. This approach helps the model maintain robustness against a broader range of attacks, preventing it from overfitting to the latest adversarial examples at the expense of forgetting previously encountered ones. This continuity in learning strengthens the model’s overall resilience by preserving knowledge from past adversarial challenges while adapting to new ones. 

MemLoss uses Memory Adversarial Examples as a new source of data without any additional computational cost for generating or costs for gathering them. This additional data source can help the model to learn more generalizable decision boundaries.

With this approach, we introduce the concept of recycling Memory Adversarial Examples for ongoing use in each epoch of adversarial training. This method can be integrated into various adversarial training frameworks to enhance model robustness while preserving high accuracy on clean data.

\subsection{Preliminaries}\label{sec:Preliminaries}
As discussed earlier, the dataset is defined as $\{(x^i, y^i)\}_{i=1}^{n}$, where $x_i$ is an input and $y_i$ is its corresponding label. For simplicity, we omit the index $i$. Now, we introduce a new notation: $x'_k$ represents the adversarial example generated at epoch $k$ of training. With this definition, the adversarial example generated in the current epoch of the training process becomes $x'_n$ which n is the number of the current epoch. We should note that $x'_n$ is the same as $x'$ in the adversarial training literature. 
\subsection{How Does Memory Help in Adversarial Training?}

The concept of MemLoss involves reusing adversarial examples from prior epochs as an additional training signal. By including these Memory Adversarial Examples in the training process, the model benefits from a form of continuity in its decision boundary adjustment, which can lead to more stable training.

We hypothesize that after training in the $n^{th}$ epoch, the network adjusts its decision boundaries in response to the adversarial examples generated in that epoch $x'_n$ and the clean input $x$. This recalibration, while adaptive, may inadvertently expose the network to vulnerabilities against Memory Adversarial Examples.

\begin{figure}[htbp] 
  \centering
  \includegraphics[width=0.3\linewidth]{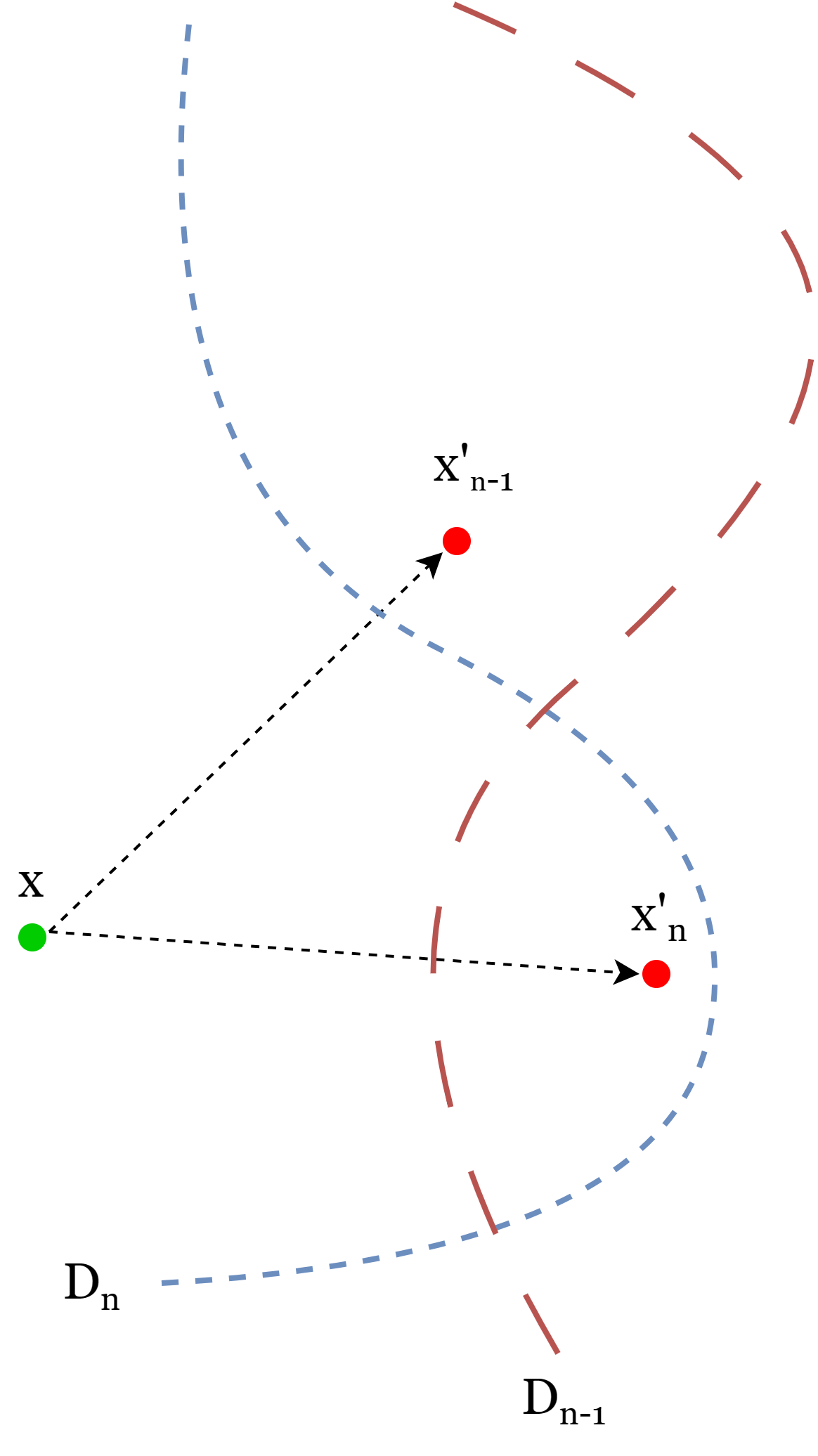}
  \caption{Post $n^{\text{th}}$ epoch, the decision boundaries may become susceptible to previous Memory Adversarial Examples.}
  \label{fig:decision_boundary}
\end{figure}

Figure \ref{fig:decision_boundary} illustrates this phenomenon, where $D_i$ symbolizes the decision boundary demarcated by the network following the $i^{th}$ epoch of adversarial training and $x'_i$ is the adversarial example generated in the $i^{th}$ epoch of adversarial training. In epoch $n-1$, the model becomes robust to the adversarial example $x'_{n-1}$, depicted as a red point and adjusts its decision boundary accordingly. However, in the subsequent epoch, the model may forget this information as it focuses on the new adversarial example $x'_n$, leading to potential vulnerabilities. This figure shows how the decision boundary ($D_{n-1}$) shifts from one epoch to another, which might leave the model susceptible to adversarial examples from previous epochs. 

This phenomenon, characterized as the network's 'forgetting' of the Memory Adversarial Example, underscores a critical vulnerability in the dynamic adaptation of decision boundaries, revealing a nuanced interplay between learning efficiency and robustness in the face of evolving adversarial threats.

By utilizing Memory Adversarial Examples, MemLoss prevents the model from forgetting past adversarial directions, thereby enhancing robustness without compromising clean accuracy. This consistent reinforcement helps the model create smoother and more generalizable decision boundaries, contributing to improved accuracy even on clean inputs. Additionally, we know that the diversity of attacks used in the adversarial training can help it to improve its robustness \cite{goodfellow2014explaining}. The Memory Adversarial Examples can be considered as an additional source of attack in the process of adversarial training, thus helping it to have a more diverse set of adversarial attacks. This explains why our experiments have shown that MemLoss can achieve higher natural accuracy compared to traditional adversarial training methods like TRADES and MART.

\subsection{Toy Problem}
\begin{figure}[htbp]
    \centering
    \includegraphics[width=0.5\linewidth]{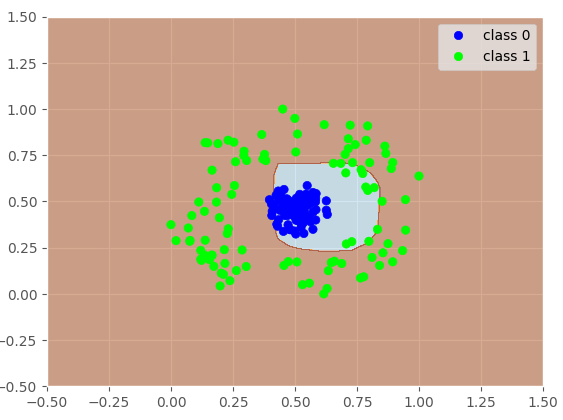}
    \caption{Decision boundary on clean data after epoch 116}
    \label{fig:decision_boundary_clean}
\end{figure}

To explore the impact of adversarial training on model robustness and its behavior against Memory Adversarial Examples, we studied a toy setting involving a 2D binary classification problem. The dataset includes two classes: blue (class 0) and green (class 1) points, as shown in Fig \ref{fig:decision_boundary_clean}. We trained a two-layer multilayer perceptron (MLP) with ReLU activation in each layer and a sigmoid activation function at the output. We used binary cross-entropy loss with an SGD optimizer (learning rate $0.005$, momentum $0.9$, weight decay $0.0005$). The model was adversarially trained with the Projected Gradient Descent (PGD) attack, using $\epsilon=0.1$ in the $l_{\infty}$ norm, until convergence (120 epochs).

\begin{figure}[htbp]
    \centering
    \begin{subfigure}[b]{0.4\linewidth}
        \centering
        \includegraphics[width=\textwidth]{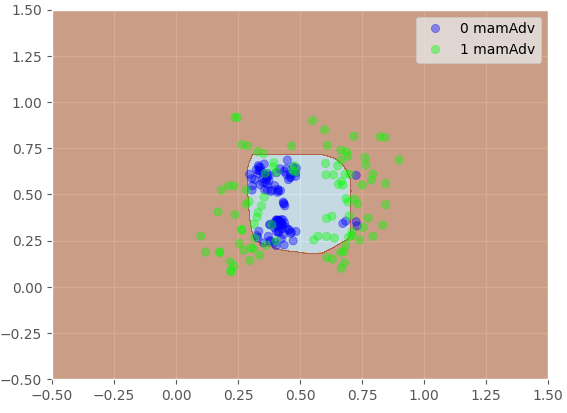}
        \caption{The decision boundary of epoch 115 training on adversarial examples generated in epoch 115.}
        \label{fig:memAdv_before}
    \end{subfigure}
    \hfill
    \begin{subfigure}[b]{0.4\linewidth}
        \centering
        \includegraphics[width=\textwidth]{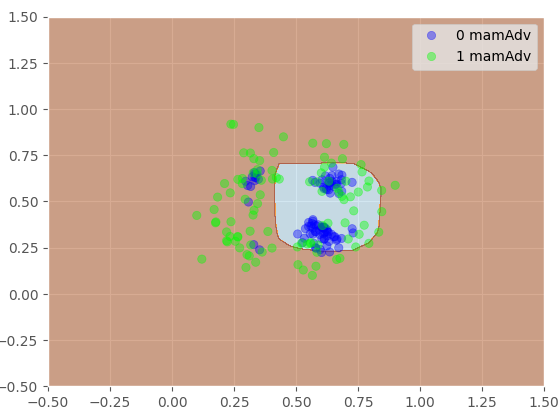}
        \caption{The decision boundary of epoch 116 training on adversarial examples generated in epoch 116.}
        \label{fig:memAdv_epoch_117}
    \end{subfigure}
    \hfill
    \begin{subfigure}[b]{0.4\linewidth}
        \centering
        \includegraphics[width=\textwidth]{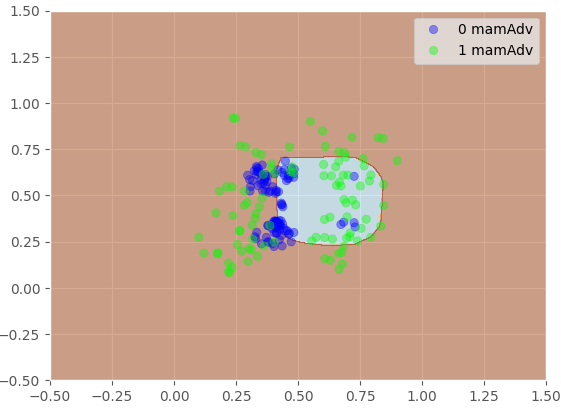}
        \caption{The decision boundary of epoch 116 training on adversarial examples generated in epoch 115.}
        \label{fig:memAdv_after}
    \end{subfigure}
    
    \caption{In this figure, we plot the decision boundary of epoch 116 on adversarial examples generated in both epochs 115 and 116. We also show the decision boundary of epoch 115 on adversarial examples generated in epoch 115. This illustrates how newly generated adversarial examples alter the decision boundary, resulting in increased error on older adversarial examples.}
    \label{fig:decision_boundary_memAdv}
\end{figure}

Fig \ref{fig:decision_boundary_clean} shows the decision boundary on clean data after training in epoch 116, where the orange region corresponds to the predicted class for green, and the sky blue region corresponds to the predicted class for blue. Fig \ref{fig:decision_boundary_memAdv}  illustrates that during training in epoch 116, the model focuses on adversarial examples generated in the current epoch, while neglecting the  Memory Adversarial Examples from previous epochs. In Figure \ref{fig:memAdv_before}, we present adversarial examples generated in epoch 115 along with the corresponding decision boundary for epoch 115. Figure \ref{fig:memAdv_epoch_117} shows adversarial examples generated for epoch 116 and the decision boundary for epoch 116. Figure \ref{fig:memAdv_after} depicts adversarial examples generated in epoch 115 with the decision boundary from epoch 116 (See Fig \ref{fig:memAdv_after} and Fig \ref{fig:memAdv_before}). The decision boundary of epoch 116 is biased towards the adversarial examples from that epoch, leading to the forgetting of adversarial examples from epoch 115. This observation motivates the need to regularize adversarial training with Memory Adversarial Examples to mitigate this effect and maintain robustness across epochs.


\subsection{MemLoss}
Our approach focuses on recycling adversarial examples from multiple previous epochs. Inspired by the concept of momentum \cite{rumelhart1986learning}, which combines previous and current directions in the update step, we integrate multiple Memory Adversarial Examples into the current training epoch.

Building upon the foundation of the TRADES method \cite{zhang2019theoretically}, we enhance the adversarial training process by incorporating a new loss term, named \textit{MemLoss}. This additional term is designed to utilize Memory Adversarial Examples from the past \( K \) iterations, enriching the training procedure with a wider spectrum of adversarial scenarios. The formal definition of our proposed loss function is as follows:

\begin{equation}
\begin{split}
    \mathcal{L}_{\text{Total}}(\theta, x, y, n, K) = & \ \mathcal{L}_{\text{TRADES}}(\theta, x, y) \\
    & + \mathcal{L}_{\text{MemLoss-K}}(\theta, x, y, n, K)
\end{split}
\end{equation}

Here, \( \mathcal{L}_{\text{TRADES}} \) represents the loss function as defined in TRADES, and \( \mathcal{L}_{\text{MemLoss-K}} \) is our newly introduced term that leverages Memory Adversarial Examples from the previous \( K \) epochs. \( n \) is the number of the current epoch, and K is the number of previous epochs that we want to use Memory Adversarial Examples generated in those epochs.

Similar to TRADES, adversarial examples are generated by solving the following optimization problem:

\begin{equation}
    x' = \arg \max_{x'} KL(f_{\theta}(x) \| f_{\theta}(x'))
\end{equation}

We use the Projected Gradient Descent (PGD) method to solve this problem iteratively. Additionally, we save adversarial examples generated in each iteration for use in subsequent iterations. We formally define \( \mathcal{L}_{\text{MemLoss-K}} \) as:

\begin{equation} \label{eq:mem-k}
    \begin{aligned}
    \mathcal{L}_{\text{MemLoss-K}}(\theta, x, y, n, K) = & \sum_{k=n-K}^{n-1} \beta'_k \cdot KL(f_{\theta}(x) \| f_{\theta}(x'_k)) \\
    & \cdot (1 - p_y(x'_k, \theta))
    \end{aligned}
\end{equation}

In this equation:

\begin{itemize}
    \item \( x'_k \) denotes the adversarial example from iteration \( k \).
    \item \( K \) is the total number of previous epochs that we are recycling adversarial examples from.
    \item \( \beta'_k \) are weighting coefficients that can be tuned to control the influence of each previous adversarial example.
    \item \( p_y(x'_k, \theta) \) represents the model's predicted probability for the correct class \( y \) on adversarial example \( x'_k \).
\end{itemize}

The introduction of \( \mathcal{L}_{\text{MemLoss}} \) aims to balance the network's adaptation to current threats while retaining resilience against past adversarial patterns. The KL divergence between the model's predictions for the clean input \( x \) and each Memory Adversarial Example \( x'_k \) encourages the model to align its predictions for \( x'_k \) more closely with those for \( x \). 
On the other hand, the model may not be robust against some adversarial examples generated in previous steps. Therefore, these examples need to be given more attention and emphasis. For this purpose, as shown in the equation \ref{eq:mem-k}, we define a weighting term.
This weighting term is grounded in the intuition that if $x'_k$ is not correctly classified after the previous
epochs (meaning it was a "hard" adversarial example), then it should be given more emphasis in the
current training epoch, and if the model is already classifying $x'_k$ correctly (implying it was an "easy" adversarial
example), this term assigns a smaller weight, indicating less emphasis is needed in the training
process.

By incorporating Memory Adversarial Examples, \( \mathcal{L}_{\text{MemLoss}} \) not only encourages the model to retain robustness against a broader range of adversarial examples but also dynamically adjusts the influence of these examples based on their classification difficulty. This nuanced approach aims to improve the overall effectiveness of adversarial training.

Our method is orthogonal to other adversarial training frameworks, meaning it can be seamlessly integrated with various techniques. Our experimental results demonstrate that adding MemLoss to both TRADES and HAT adversarial training methods improves their performance, highlighting the versatility and effectiveness of our approach. The MemLoss term can be added to any adversarial loss function as follows:

\begin{equation}
    \begin{aligned}
    \mathcal{L}_{\text{Total}}(\theta, x, y, n, K) = & \, \mathcal{L}(\theta, x, y) \\
    & \, +  \mathcal{L}_{\text{MemLoss-K}}(\theta, x, y, n, K)
    \end{aligned}
\end{equation}

Although the form of the MemLoss regularizer shares some conceptual similarities with those of TRADES \cite{zhang2019theoretically} and MART \cite{wang2019improving}, the critical difference lies in the use of 'memory' which ensures that adversarial training benefits from previously generated adversarial directions, enhancing stability and robustness in ways that existing methods do not.

\begin{algorithm}
\caption{Adversarial Training with Memory of \( K \) Previous Adversarial Examples}
\label{algorithm:memloss}
\begin{algorithmic}[1]
\small
\STATE {\bfseries Note:} \( x^{i}_k \) represents the adversarial example for sample \( x^{i} \) generated in epoch \( k \), and \( n \) is the current training epoch.
\STATE {\bfseries Input:} Training dataset \( S = \{(x^{i}, y^{i})\}_{i=1}^n \)
\STATE {\bfseries Parameters:} Batch size \( m \), learning rate \( \eta \), weight of robust loss \( \beta \), weights of MemLoss \( \beta'_k \), attack radius \( \epsilon \), attack step size \( \alpha \), number of attack iterations \( T \), number of previous epochs \( K \)
\STATE Initialize network parameters \( \theta \)
\REPEAT
    \STATE Sample mini-batch \( \{(x^{i_j}, y^{i_j})\}_{j=1}^m \) from \( S \)
    \FOR{\( j = 1 \) \textbf{to} \( m \)}
        \STATE Initialize \( x'^{i_j}_n \leftarrow x^{i_j} + 0.001 \cdot \mathcal{N}(0, I) \)
        \FOR{\( t = 1 \) \textbf{to} \( T \)}
            \STATE \( x'^{i_j}_n \leftarrow \Pi_{B(x^{(i_j)}, \epsilon)}\big( x'^{i_j} + \alpha \cdot \text{sign}\big( \nabla_{x'} KL( f_{\theta}(x^{(i_j)}) \| f_{\theta}(x'^{i_j}) ) \big) \big) \)
        \ENDFOR
        \IF{\( n < K \)}
            \FOR{\( t = K \) \textbf{down to} \( n+1 \)}
                \STATE \( x'^{i_j}_t \leftarrow x^{i_j} \)
            \ENDFOR
        \ENDIF
    \ENDFOR
    \STATE Update network parameters:
    \STATE \( \theta \leftarrow \theta - \frac{\eta}{m} \sum_{j=1}^m \nabla_{\theta} \Big( CE( y^{i_j}, f_{\theta}(x^{i_j}) ) + \beta \cdot KL( f_{\theta}(x^{i_j}) \| f_{\theta}(x'^{i_j}_n) )  
    + \sum_{k=1}^{K} \beta'_k \cdot KL( f_{\theta}(x^{i_j}) \| f_{\theta}(x'^{i_j}_k) ) \cdot ( 1 - p_y( x'^{i_j}_k, \theta ) ) \Big) \)
    \FOR{\( j = 1 \) \textbf{to} \( m \)}
        \STATE Update memory adversarial examples for \( x^{(i_j)} \):
        \FOR{\( k = K \) \textbf{down to} \( 2 \)}
            \STATE \( x'^{i_j}_{k-1} \leftarrow x'^{i_j}_{k} \)
        \ENDFOR
        \STATE \( x'^{i_j}_K \leftarrow x'^{i_j}_n \)
    \ENDFOR
\UNTIL{training converged}
\normalsize
\end{algorithmic}
\end{algorithm}

\textbf{MemLoss algorithm:} The pseudo-code for MemLoss adversarial training is demonstrated in algorithm \ref{algorithm:memloss}. In the epochs that are less than K, we use the original examples memory adversarial examples for Memory Adversarial Examples that we don't have, thus the terms related to these Memory Adversarial Examples become zero. In each epoch, our algorithm will save the adversarial examples generated to use in the next epochs as Memory Adversarial Examples.

\section{Experiments}\label{sec:Experiments}
This section presents the evaluation of MemLoss. First, we test our method using the ResNet~\cite{resnet} architecture and study its performance on different datasets and against various adversarial attacks (see Section~\ref{sec:performance_evaluation}). Then, we examine how our method performs under other threat models (see Section~\ref{sec:other_threat_models}). Finally, we conduct experiments to analyze MemLoss in detail (see Section~\ref{sec:analyze}).

\subsection{Performance Evaluation}\label{sec:performance_evaluation}
In our investigation, we evaluate the efficacy of MemLoss, employing the ResNet-18 architecture \cite{resnet} across three distinct datasets: CIFAR-10, CIFAR-100 \cite{krizhevsky2009learning}, and SVHN \cite{netzer2011reading}. This evaluation entailed a comparative analysis of MemLoss against prominent adversarial defense strategies: AT \cite{mkadry2018towards}, TRADES \cite{zhang2019theoretically}, HAT \cite{rade2022reducing} and MART \cite{wang2019improving}.


\subsubsection{Training Setup}
For training, we used an SGD optimizer with Nesterov momentum. The learning rate strategy comprised cyclic learning rates \cite{smith2018superconvergence} with cosine annealing, setting a maximum learning rate of 0.21 for CIFAR-10 and CIFAR-100, and 0.05 for SVHN. Training duration was set to 70 epochs for CIFAR-10 and CIFAR-100, and 22 for SVHN. For $\ell_{\infty}$ training, we applied a PGD attack with a maximum perturbation $\epsilon = 8/255$, over $T = 10$ iterations for all datasets. The PGD step size was $\alpha =\epsilon/4 = 2/255$ for CIFAR-10 and CIFAR-100, and $\alpha = 1/255$ for SVHN. For MemLoss (MemLoss term plus TRADES) configurations, we set $\beta' = 2$, and $\beta = 5$ for all datasets. For MemLoss + HAT (HAT plus MemLoss term) we set $\beta' = 1$, $\beta = 2$ and 
$\gamma = 0.5$ for all datasets. TRADES employed a regularization parameter $\beta$ of 5.0 for CIFAR-10 and SVHN, and 6.0 for CIFAR-100, whereas MART used a consistent $\beta = 5.0$. For HAT we use configuration based on their paper, we set $\gamma = 0.5$ for all datasets, for CIFAR-10 and SVHN we set $\beta = 2.5$ and for CIFAR-100 we set $\beta = 3.5$. More details can be found in the supplementary. We assume \( K = 1 \), as experiments indicate that considering only the adversarial example generated in the previous step yields the best results.

Here, we focus on $K=1$, and simplify $\beta'_{n-1}$ to $\beta'$ and $x'_{n-1}$ to $x''$. We choose $K=1$ for two main reasons. First, as $K$ increases, we will need more RAM and computation to save and use Memory Adversarial Examples. Second, based on our empirical experiments, which can be seen in the supplementary section, $K=1$ worked better than other values for $K$. Overall, $K>1$ is out of the scope of this paper and can be studied in future works.   

\subsubsection{Evaluation Setup}
The robustness of models was evaluated using early stopping \cite{rice2020overfitting} to identify the model with maximal resilience against a PGD attack with $T = 20$ iterations. Additionally, the robust accuracy of our models was assessed using AutoAttack \cite{AA}, which integrates an ensemble of four distinct attack strategies, including a black-box approach. We also did some experiments to check the gradient obfuscation problem in our method, which can be found in the supplementary section, and there was no sign of this problem considering our method. 

\subsubsection{Results}
The results show that incorporating Memory Adversarial Examples through MemLoss leads to significant improvements in both clean accuracy and robustness across different datasets compared to existing adversarial training methods. Table \ref{tab:table1} presents the performance of MemLoss combined with Trades and MemLoss combined with HAT, compared to other approaches like TRADES, MART, and HAT. On CIFAR-10, MemLoss with Trades achieves a robust accuracy of 49.79\%, outperforming all other methods, while maintaining competitive clean accuracy. In CIFAR-100, MemLoss also improves robustness to 24.24\%, higher than TRADES and HAT, while increasing clean accuracy to 56.81\%. Furthermore, on SVHN, MemLoss achieves the highest robustness at 51.58\%, without significantly compromising clean accuracy. Combining MemLoss with HAT further enhances results, achieving the best clean accuracy across all datasets, highlighting the effectiveness of Memory Adversarial Examples in improving both natural and adversarial performance.

\begin{table}[ht] 
\centering
\resizebox{0.7\textwidth}{!}{ 
\begin{tabular}{lcccccc}
\hline
\textbf{Method} & \multicolumn{2}{c}{\textbf{CIFAR-10}} & \multicolumn{2}{c}{\textbf{CIFAR-100}} & \multicolumn{2}{c}{\textbf{SVHN}} \\
& Clean & Robust & Clean & Robust & Clean & Robust \\ 
\hline
Standard & 94.52 & 0.0 & 74.15 & 0.0 & 95.72 & 0.10 \\
\hdashline
AT & 84.16 & 45.83 & 54.94 & 21.04 & \textbf{92.23} & 41.76 \\
TRADES & 82.85 & 48.84 & 54.26 & 21.98 & 90.49 & 49.65 \\
MART & 79.86 & 47.22 & 50.21 & 22.78 & 88.42 & 44.86 \\
HAT & 84.35 & 47.41 & 55.99 & 21.74 & 91.79 & 51.13 \\
MemLoss + Trades & 83.08 & \textbf{49.79} & 56.81 & \textbf{24.24} & 90.10 & \textbf{51.58} \\
MemLoss + HAT & \textbf{85.51} & 49.24 & \textbf{59.84} & 22.76 & 91.40 & 51.71 \\
\hline
\end{tabular}
}
\caption{Comparison of MemLoss with other adversarial defenses in $\ell_{\infty}$ perturbation scenario with maximum size of perturbation $\epsilon=8/255$. We used ResNet-18 and reported Clean accuracy and robust accuracy (using AutoAttack) on CIFAR-10, CIFAR-100, and SVHN.}
\label{tab:table1}
\end{table}

In Table\ref{tab:table22} we can see the comparison of MemLoss with other methods against other adversarial attacks on CIFAR-10. The PGD+ attack used in this table is the PGD attack with 40 iterations and 5 restarts \cite{carmon2019unlabeled} and the CW attack is the PGD attack with CW loss \cite{carlini2017towards} and 40 iterations. More information can be found in the supplementary section. 

The results highlight the effectiveness of MemLoss in maintaining a competitive level of adversarial robustness across various attack types. Notably, MemLoss achieves the highest adversarial accuracy under PGD+ and CW attacks, outperforming TRADES, MART, and HAT. Furthermore, MemLoss + HAT achieves the best clean accuracy, showcasing the strength of our approach in balancing clean accuracy with adversarial robustness.

\begin{table}[ht] 
\centering
\resizebox{0.7\textwidth}{!}{ 
\begin{tabular}{lcccc}
\hline
\textbf{Method} & \textbf{Clean} & \textbf{AutoAttack} & \textbf{PGD+} & \textbf{CW} \\
\hline
Standard & 94.52 & 0.0 & 0.0 & 0.0 \\
\hdashline
AT & 84.16 & 45.83 & 48.17 & 48.36 \\
TRADES & 82.85 & 48.84 & 51.72 & 49.90 \\
MART & 79.86 & 47.22 & 52.71 & 48.84 \\
HAT & 84.35 & 47.41 & 49.93 & 48.65 \\
MemLoss + Trades & 83.08 & \textbf{49.79} & \textbf{52.97} & \textbf{50.89} \\
MemLoss + HAT & \textbf{85.51} & 49.24 & 52.22 & 50.59 \\
\hline
\end{tabular}
}
\caption{Comparison of MemLoss using ResNet-18 on CIFAR-10 against other adversarial attacks with adversarial defense methods in $\ell_{\infty}$ perturbation scenario.}
\label{tab:table22}
\end{table}

Overall, MemLoss demonstrates minimal or no reduction in adversarial robustness compared to other state-of-the-art methods, even under more challenging attacks. These results further validate the efficacy of incorporating Memory Adversarial Examples into adversarial training to improve the robustness of models while preserving clean accuracy.

\subsection{Other Threat Models}\label{sec:other_threat_models}
In our investigation of alternative threat models, we also conduct experiments with different attack configurations by evaluating our method against two other adversaries on CIFAR-10: $\ell_\infty$ with $\epsilon=12/255$ and $\ell_2$ with $\epsilon=128/255$. As shown in Table\ref{tab:table4}, our method demonstrated superior performance over TRADES, notably improving both clean accuracy and robustness.

\begin{table}[ht] 
\centering
\resizebox{0.7\textwidth}{!}{ 
\begin{tabular}{lcccc}
\hline
\textbf{Norm} & $\epsilon$ & \textbf{Method} & \textbf{Clean} & \textbf{Robust} \\
\hline
\multirow{2}{*}{$\ell_{\infty}$} & \multirow{2}{*}{12/255} & TRADES & 73.35 & 32.77 \\
& & MemLoss + Trades &  \textbf{76.34} &  \textbf{33.10} \\
\multirow{2}{*}{$\ell_{2}$} & \multirow{2}{*}{128/255} & TRADES & 87.41 & 68.99 \\
& & MemLoss + Trades & \textbf{88.84} & \textbf{69.14}  \\
\hline
\end{tabular}
}
\caption{Comparison of MemLoss and TRADES under $\ell_p$ constrained adversaries, using ResNet-18 on CIFAR-10.}
\label{tab:table4}
\end{table}

\subsection{Analyzing parameter $\beta$}
\label{sec:analyze}
\begin{figure}[H]
  \centering
  \includegraphics[width=0.6\linewidth]{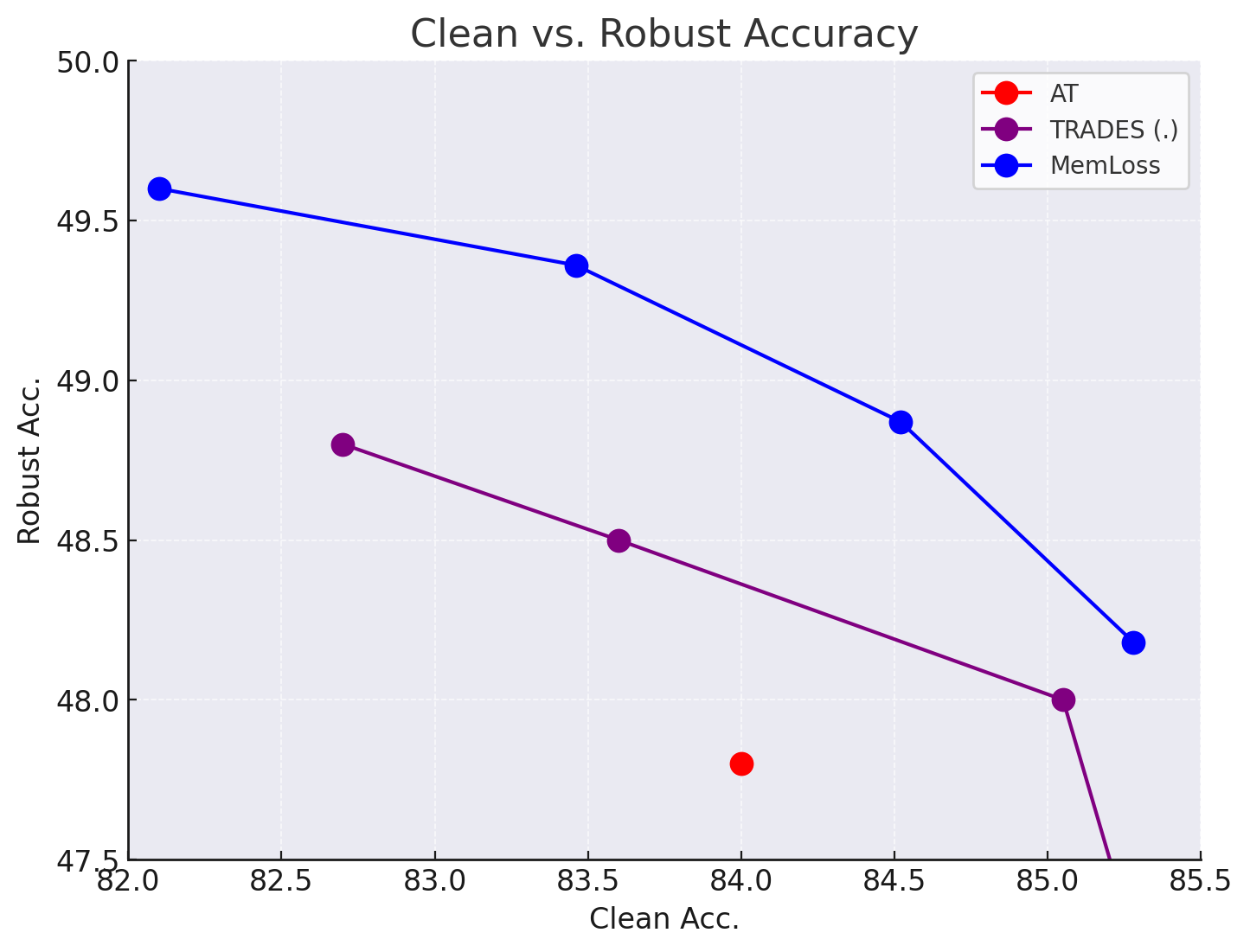}
   \caption{Comparison of accuracy and robustness trade-off between MemLoss, TRADES, and AT.}
   \label{fig1}
\end{figure}

For analyzing the impact of adding MemLoss to the TRADES, we examined the effect of parameter $\beta$ on the performance of TRADES and TRADES + MemLoss. 
Figure \ref{fig1} shows the impact of the parameter $\beta$ on both MemLoss + TRADES and TRADES, emphasizing the trade-off between clean accuracy and robustness. In this experiment, for MemLoss, $\beta'$ is fixed at 2, while $\beta$ increases from 2.5 to 5 from left to right. For TRADES, $\beta$ decreases from 4 to 1.5. As seen in the plot, MemLoss consistently maintains higher robustness at similar levels of clean accuracy compared to TRADES. This suggests that MemLoss is more effective at balancing the trade-off between robustness and accuracy, allowing the model to achieve better results in both metrics simultaneously. The figure also highlights that MemLoss efficiently leverages $\beta$ to navigate the robustness-accuracy trade-off, making it more adaptable than TRADES. Overall, MemLoss demonstrates its capability to enhance robustness without compromising clean accuracy, which is a highly desirable property for adversarial training.

\section{Conclusion}
\label{sec:Conclusion}
In this paper, we introduced Memory Adversarial Examples, a novel concept that leverages adversarial examples generated in previous training epochs to enhance robustness and accuracy in adversarial training. Our method, MemLoss, effectively uses these historical adversarial examples to strengthen model defenses against both previously encountered and new adversarial threats.

Experimental results on benchmark datasets show that MemLoss consistently outperforms traditional adversarial training methods, such as TRADES and MART, in key metrics. By incorporating Memory Adversarial Examples, our method improves model robustness while maintaining or even enhancing clean data accuracy, effectively addressing the common trade-off in adversarial training. The results highlight the benefit of integrating memory into the adversarial training process to achieve a better balance between clean and robust performance.

MemLoss is a promising direction for adversarial training, with potential for further exploration. Future work includes optimizing Memory Adversarial Examples selection strategies, efficiently incorporating historical adversarial knowledge, and evaluating MemLoss across more diverse datasets and deeper model architectures. These efforts could provide broader insights and lead to further advancements in robust machine learning.

Despite its promising results, MemLoss has limitations. Future research should investigate its generalizability across different model architectures and evaluate the impact of augmenting the training process with additional synthesized data. Addressing these limitations could further enhance the effectiveness and versatility of adversarial robustness solutions.

\section*{Data and Code Availability}

The code and data used in this study are publicly available at: \url{https://github.com/soroush-mim/hat}.

\section*{Declaration of generative AI and AI-assisted technologies in the writing process}

During the preparation of this work, the author(s) used ChatGPT for grammar checking and paraphrasing. After using this tool, the author(s) reviewed and edited the content as needed and take(s) full responsibility for the content of the publication.
\bibliographystyle{elsarticle-num}  
\bibliography{sample}

\clearpage
\setcounter{page}{1}
\appendix
\section{Supplementary Material}
\subsection{Experimental Setting}
The particulars of our experimental procedures, encompassing training and evaluation protocols, are cataloged in this section.

\subsubsection{Training setup.}
Our experiment setup is very much the same as the HAT\cite{rade2022reducing} paper.
For the CIFAR-10 and CIFAR-100 datasets \cite{krizhevsky2009learning}, our experiments are built upon the ResNet-18 architecture \cite{resnet}, while for SVHN \cite{netzer2011reading}, we utilize the PreAct ResNet-18 variant. The optimization process is facilitated by the Stochastic Gradient Descent (SGD) optimizer with Nesterov accelerated gradient, a momentum of 0.9, and weight decay set to 0.0005. Cyclic learning rates with cosine annealing \cite{smith2018superconvergence} are employed, with the peak learning rate fixed at 0.21 for CIFAR-10 and CIFAR-100, and 0.05 for SVHN. The models for CIFAR-10 and CIFAR-100 undergo a training regime of 70 epochs with a batch size of 128, while for SVHN, the training duration is abbreviated to 22 epochs.

\subsubsection{Adversarial example computation.}
The generation of adversarial examples during training adheres to the $\ell_{\infty}$-PGD with the following parameters: $\ell_{\infty}$ norm $\epsilon = 8/255$, step size $\alpha = 2/255$, and the attack iterated for $T = 10$ rounds. TRADES, MART, and HAT's hyperparameters are fine-tuned as follows: for HAT, we fix $\gamma = 0.5$ and employ $\beta = 2.5$ for CIFAR-10 and SVHN, and $\beta = 3.5$ for CIFAR-100. The TRADES regularization parameter $\beta$ is calibrated to 5.0. For MEMLoss configurations, we set $\beta' = 2$, and $\beta = 5$, and for MemLoss + HAT we set $\beta' = 1$, $\beta = 2$, and $\gamma = 0.5$for all datasets. 

\subsection{Evaluation Protocol}
Throughout our training process, we implement an early stopping mechanism. This involves closely monitoring the model's robustness against a PGD attack (with $T = 20$ iterations) on the test set and selecting the best-performing snapshot for further assessments. To evaluate the $\ell_{\infty}$ robustness, our models are subjected to AutoAttack (AA) \cite{AA}, utilizing the standard implementation available at \url{https://github.com/fra31/auto-attack}.

\subsubsection{Setup for Other Attack Configurations}
The training setup employed for evaluating under varying threat models, specifically on CIFAR-10, mirrors the one detailed earlier. For $\ell_{2}$ perturbations, our training adversary parameters are set as follows: $\epsilon = 128/255$, $\alpha = 15/255$, and $K = 10$. We select $\beta = 3$ and $\beta' = 2$ for MemLoss, and $\beta = 5.0$ for TRADES. For $\ell_{\infty}$ perturbations with norm $\epsilon = 12/255$, we employ $\alpha = 4/255$ and maintain $T = 10$. Additionally, for MemLoss, $\beta = 2.5$ and $\beta' = 2$ are chosen, while $\beta = 6.0$ is used for TRADES.

\subsubsection{Evaluation with Other Adversaries}
Beyond the specified threat models, we examine the resilience of our robust models trained on CIFAR-10 against more traditional adversaries. We benchmark the $\ell_{\infty}$ robustness using PGD with $\epsilon = 8/255$, $\alpha = 0.01$, $T = 40$ steps, and $r = 5$ restarts, in accordance with the evaluation protocol by Carmon et al. \cite{carmon2019unlabeled}, herein referred to as PGD+. Our models' susceptibility to the PGD with CW loss \cite{carlini2017towards} is also tested, utilizing the same $\epsilon$ and $\alpha$ values but over $K = 40$ steps.

\subsection{Checking Gradient Obfuscation}
For the robustness assessment of our models, we utilize AutoAttack \cite{AA}, a tool renowned for its dependable robustness evaluation capabilities. Moreover, to preclude the likelihood of gradient masking, additional verification procedures are implemented.
we scrutinize the influence of increasing both the attack iterations
T and the random restarts
r on the robustness of a MemLoss with Trades-trained model on CIFAR-10 subjected to $\ell_{\infty}$
perturbations of magnitude 
8/255 with 50 epochs. Our findings indicate that enhancements in the number of iterations 
T or random restarts 
r result in only a nominal decrease in robust accuracy, as delineated in Tables 1 and 2. This suggests that the attack is reaching its potential effectiveness and is free from issues of gradient obfuscation.

\begin{table}[ht]
\centering
\begin{tabular}{cc}
\hline
\textbf{No. of steps} \( T \) & \textbf{Robust} \\
\hline
40 & 52.23 \\
200 & 52.13 \\
500 & 52.10 \\
1000 & 52.10 \\
\hline
\end{tabular}
\caption{Robust accuracy vs. number of steps \( T \) of PGD attack on CIFAR-10}
\label{tab:steps_vs_robust}
\end{table}

\begin{table}[ht]
\centering
\begin{tabular}{cc}
\hline
\textbf{No. of restarts} \( r \) & \textbf{Robust} \\
\hline
1 & 52.23 \\
5 & 51.94 \\
10 & 51.86 \\
20 & 51.82 \\
\hline
\end{tabular}
\caption{Robust accuracy vs. number of random restarts \( r \) of PGD-40 attack on CIFAR-10}
\label{tab:restarts_vs_robust}
\end{table}

\subsection{Other Ways of Using Memory Adversarial Examples}
We discuss additional methods for incorporating memory adversarial examples into training, along with their potential impact on model performance.

The first way for this is MemLoss but without the weighting term. In this case the MemLoss term will be like this:

\begin{equation}
    \mathcal{L}_{\text{MemLoss}}(\theta, x, y) = KL(f_{\theta}(x) || f_{\theta}(x''))
\end{equation}
which we name it MemLoss Version 2 (V2).

Another way to use Memory Adversarial examples is defining MemLoss like this:
\begin{equation}
    \mathcal{L}_{\text{MemLoss}}(\theta, x, y) = KL(f_{\theta}(x'') || f_{\theta}(x'))
\end{equation}
The idea behind this term is that we assume that the network has learned robust representation for $x''$, maybe it will be beneficial to use this representation in the process of learning a robust representation for $x'$. We name this method MemLoss Version 3 (V3).
\begin{table}[ht]
\centering
\begin{tabular}{lcc}
\hline
\textbf{Method} & \textbf{Clean} & \textbf{Robust} \\
\hline
V1 & 84.52 & 48.87  \\
V2 & 83.54 & 48.60  \\
V3 & 82.19 & 48.58  \\
\hline
\end{tabular}
\caption{Comparison of different versions of MemLoss using ResNet-18 on CIFAR-10. We show the main method, introduced in our work, with V1. For V1 we set $\beta=3.0$ and $\beta'=2.0$, for V2 we set $\beta=4.0$ and $\beta'=1.0$ and for V3 we set $\beta=4.0$ and $\beta'=2.0$. We report standard accuracy and AutoAttack robustness. We can see that version 1 of MemLoss is superior both in standard accuracy and robustness to other versions.}
\label{tab:table2}
\end{table}



\end{document}